\documentstyle[twocolumn,lsalike]{article}

\newcommand{\ul}{\underline}

\newcounter{sentence}
\newcommand{\sent}[1]{\refstepcounter{sentence}%
\label{#1}%
\ref{#1}}

\newenvironment{rei}{\begin{flushleft} \setlength{\tabcolsep}{0.5\tabcolsep}}{\end{flushleft}}
\newcommand{\border}{\rule{\textwidth}{0.2mm}}
\newcommand{\halfborder}{\rule{\columnwidth}{0.2mm}}

\title{Possessive pronouns as determiners\\
  in Japanese-to-English machine translation}

\author{Francis {\sc Bond}, Kentaro {\sc Ogura}, Satoru {\sc Ikehara}
  \\ {\bf NTT Communication Science Laboratories} \\ 
  1-2356 Take, Yokosuka-shi, Kanagawa-ken, {\sc Japan} 238-03 \\ 
  Tel: 0468-59-5827; Fax: 0468-59-3633\\ 
  {\tt \{bond,ogura,ikehara\}@nttkb.ntt.jp}}

\date{PACLING~'95: April 1995\footnotemark[2]}

\begin{document}

\bibliographystyle{lsalike}


\maketitle
\begin{abstract}

  Possessive pronouns are used as determiners in English when no
  equivalent would be used in a Japanese sentence with the same
  meaning.  This paper proposes a heuristic method of generating such
  possessive pronouns even when there is no equivalent in the
  Japanese.  The method uses information about the use of possessive
  pronouns in English treated as a lexical property of nouns, in
  addition to contextual information about noun phrase referentiality
  and the subject and main verb of the sentence that the noun phrase
  appears in.  The proposed method has been implemented in NTT
  Communication Science Laboratories' Japanese-to-English machine
  translation system {\bf ALT-J/E}.  In a test set of 6,200 sentences,
  the proposed method increased the number of noun phrases with
  appropriate possessive pronouns generated, by 263 to 609, at the
  cost of generating 83 noun phrases with inappropriate possessive
  pronouns.

\end{abstract}
\renewcommand{\thefootnote}{\fnsymbol{footnote}} \footnotetext[2]{This
  paper was presented at the 2nd Pacific Association for Computational
  Linguistics Conference (PACLING~'95) and appears in the proceedings.}
\renewcommand{\thefootnote}{\arabic{footnote}}

%
%

\section{Introduction}

Possessive pronouns are often used as determiners in English when no
equivalent would be used in a Japanese sentence with the same meaning.
For example, when referring to specific family members in English, it
is normal to specify whose relations they are.  In Japanese these are
only specified if they are not obvious from the context.  For a
machine translation system to generate appropriate English when
translating from Japanese, it is necessary to determine which pronouns
should be used and when.

The similar problem of determining article usage and noun phrase
number\footnote{Japanese does not have articles, and noun phrases are
  normally not marked for number.} has recently been approached in
three ways: using expert-system-like rules to determine the
referential property and number of nouns \cite{Murata:1993a}; using
heuristic rules based on the meaning of the Japanese sentence and the
properties of the generated English to determine the referentiality
and number of English noun phrases \cite{Bond:1994} and using a
Context Monitor to maintain contextual information dynamically
\cite{Cornish:1994}.  The problem of generating possessive pronouns as
determiners in translation where there is no equivalent in the
Japanese has not previously been addressed; it requires not
only contextual information such as that used to determine noun phrase
referentiality, but also information about the conventional usage of
possessive pronouns in English.

In this paper, we propose a method of generating possessive pronouns
as determiners for noun phrases where there is no equivalent in the
Japanese, based on treating information about the conventional use of
possessive pronouns in English as a lexical property of nouns.  In
addition, the method uses contextual information about noun phrase
referentiality, the meaning and modality of the main verb and the
denotation of the subject of the sentence that the noun phrase appears
in.  The method has been implemented in NTT Communication Science
Laboratories' Japanese-to-English machine translation system {\bf
  ALT-J/E} \cite{Ikehara:1991,Ogura:1993}.

The rest of this document is organized as follows: We begin in
Section~\ref{sec:diff} by examining the distribution of possessive
pronouns in 6,200 Japanese sentences with English translations.  657
noun phrases containing possessive pronouns were found in the
translations.  Existing algorithms, described in
Section~\ref{sec:current}, are capable of translating 52\% of
these 657 noun phrases, mainly those in which there was a possessive
construction in the Japanese.  The proposed method for appropriately
generating possessive pronouns for the remaining 48\% is presented in
Section~\ref{sec:narrow}.  The result of implementing the proposed
method is evaluated in Section~\ref{sec:results}.  Finally some
concluding remarks are given in Section~\ref{sec:conc}.

\section{Differences in the use of possessive pronouns in Japanese and
  English}
\label{sec:diff}

In order to examine the use of possessive pronouns when translating
from Japanese to English, a study was made of 6,200 sentence pairs of
Japanese sentences with English translations, produced by a
professional translator.  These pairs make up a test set (taken mainly
from written Japanese such as in newspaper articles) designed to test
the capabilities of Japanese-to-English machine translation systems.
A description of the test set and it's design is given in
\namecite{Ikehara:1994}.  The use of possessive pronouns is not one of
the criteria specifically tested by the test set.


The English translations of the test set contain 657 noun phrases with
possessive pronouns.  The sentences containing these noun phrases were
examined in order to determine how the possessive pronouns could be
generated by a machine translation system.  The noun phrases were
divided into three groups, according to whether the possessive pronoun
had an equivalent in the original Japanese, or could be predicted as
an obligatory part of an English expression or if neither of the above
conditions held.

There were 193 noun phrases (30\%) in the first group (I) where the
original Japanese noun phrase contains a possessive expression, either
a pronoun or the reflexive {\it jibun\/} `self' followed by the
genitive postposition {\it no\/} `of' that indicates possession.  The
genitive pronoun construction ({\sc pronoun}-{\it no\/}) can be
directly translated into English as a possessive pronoun.  An example
is shown in sentence~(\ref{sent:explicit})\footnote{Examples are given
  with the (romanized) Japanese original, a gloss and the human
  translation.  The examples have been simplified to exemplify points
  more clearly; a new translation has been made for each simplified
  sentence.  Japanese particles are glossed as follows: {\sc top} for
  {\it wa\/} which marks the topic, {\sc obj} for {\it o\/} which
  marks the object and {\sc gen} for {\it no\/} which shows a genitive
  relation.}.

\begin{rei} 
\begin{tabular}{clllll}
(\sent{sent:explicit})
     & Jap:   &  \it kanojo-wa &  \it kare-no & \it kao-o & \it mita\\
    & Gloss: &   she-{\sc top} &  he-{\sc gen} &  face-{\sc obj} &  saw  \\
    & Eng:   & \multicolumn{4}{l}{`She saw his face'}
\end{tabular}
\end{rei}

The genitive reflexive construction {\it jibun-no\/} `one's own'
appears in 25 cases (4\% of the total).  {\it jibun-no\/} is
translated as {\sc possessive pronoun} {\it own}.  The relation
between the pronoun and its antecedent is not given explicitly, it
depends on the context.  An example of this is shown in
sentence~(\ref{sent:jibun}).  When the subject is {\it kanojo\/} `she'
{\it jibun-no\/} `one's own' is translated as {\it her own}.  If the
subject were changed to {\it kare\/} `he' or {\it John\/} then {\it
  jibun-no\/} `one's own' would be translated as {\it his own}.

\begin{rei} 
\begin{tabular}{clllll}
(\sent{sent:jibun})
& Jap:   &  \it kanojo-wa & \it  jibun-no & \it namae-o & \it wasureta\\
& Gloss: &   she-{\sc top} &  self-{\sc gen} &  name-{\sc obj} &  forgot \\
& Eng:   & \multicolumn{4}{l}{`She forgot her own name'}
\end{tabular}
\end{rei}

The second group (II), with 105 noun phrases (16\%), consisted of those in
which the possessive pronoun appeared as part of an English expression
where both the use of a possessive pronoun and its antecedent could be
deduced from the form of the expression, but there was no equivalent
possessive construction in the Japanese original sentence.  For
example, in the expression {\it 20-dai-no-josei\/} `20 generation {\sc
  gen} woman' {\it a women in her twenties\/} the possessive pronoun
{\it her\/} is an obligatory part of the English expression, and its
antecedent is always the modificant of the prepositional phrase.  An
expression may be based on a verb, as in sentence~(\ref{sent:idiom})
where the Japanese idiom {\it chie-o shiboru\/} `wring knowledge'
is translated into an English idiom {\it rack {\sc possessive pronoun}
  brains\/} in which the antecedent of the possessive pronoun is the
noun phrase that is the subject of the verbal idiom.

\begin{rei}
\begin{tabular}{cllll}
(\sent{sent:idiom})
& Jap:   &  \it kanojo-wa & \it chie-o & \it shibotta \\
& Gloss: &   she-{\sc top}  &  knowledge-{\sc obj}    &  wrung \\
& Eng:   & \multicolumn{3}{l}{`She racked her brains'}
\end{tabular}
\end{rei}

Groups I and II can be translated straightforwardly by a machine
translation system.  A discussion of how this is done in {\bf ALT-J/E}
is given in Section~\ref{sec:current}.  

\begin{table*}[htbp]
  \border
  \begin{center}
    \leavevmode
    \begin{tabular}{clcc}
      \multicolumn{2}{c}{Noun phrase type:}  & Number of & Percentage of  \\ 
      \multicolumn{2}{c}{(657 noun phrases)} & occurrences & occurrences \\ \hline
      I & Possessive expression in the original Japanese & 193 & 30\% \\
      II & English expression requires possessive pronoun & 105 & 16\% \\
      III & Noun-triggered possessive pronoun & 359 & \ul{54\%} 
    \end{tabular}
  \end{center}
  \caption{Distribution of possessive pronouns}
  \label{tab:total}
  \border
\end{table*}

The third and final group (III) consists of 359 noun phrases (54\%)
where the original Japanese had neither a possessive construction, as
in group I, nor arose in an English expression in which it was
obligatory, as in group II\@.  These noun phrases were those where
English conventionally uses a possessive pronoun to indicate a
relationship such as ownership, as in {\it my wallet}, or a family
relationship, {\it my father}, but Japanese does not.  The use of
possessive pronouns with the nouns which head the noun phrases in
group III seems to be tied to the particular words.  In particular,
words which denote {\sc body parts, work, personal possessions,
  attributes} and relational nouns such as {\sc kin} and {\sc people
  defined by their relation to another person} (e.g. {\it assailant,
  partner, subordinate\/}) were commonly translated with possessive
pronouns.  The semantic hierarchy of 2,800 categories used in {\bf
  ALT-J/E} was not fine-grained enough to identify the words by their
denotation alone.  We therefore identified the nouns manually and
marked them with a special flag in the lexicon.  These nouns will be
referred to as `trigger-nouns' as they trigger the use of possessive
pronouns when they are used in English.  We are investigating
automating the identification process using a parsed bilingual corpus
aligned at the noun phrase level.

There were 205 different trigger-nouns in the test set used.  They
headed 825 noun phrases.  The human translations of 355 out of the 825
noun phrases headed by trigger-nouns (43\%) contained possessive
pronouns, even though the original Japanese did not contain a
possessive construction, and the possessive pronoun was not part of an
English expression in which it was obligatory.  A new heuristic method
for translating these cases that uses the head noun's lexical
information as a trigger to generate possessive pronouns, in
conjunction with contextual information is proposed in
Section~\ref{sec:narrow}.

The distribution of the above three groups is summarized in
Table~\ref{tab:total}.

\section{Existing translation algorithms}
\label{sec:current}

This section describes the overall process of translation in {\bf
  ALT-J/E}, and in particular how the possessive pronouns in noun
phrases from groups (I) and (II) are translated.

The overall process of translation can be divided into seven parts.
First, {\bf ALT-J/E} splits the Japanese text into morphemes.  Second,
it analyses the sentence syntactically, often giving multiple possible
interpretations.  Third, it rewrites complicated Japanese expressions
into simpler ones.  Fourth, {\bf ALT-J/E} semantically evaluates the
various interpretations.  Fifth, syntactic and semantic criteria are
used to select the best interpretation.  Sixth, the selected
interpretation is transferred into English.  Finally, the English
sentence is adjusted to give the correct inflectional forms.

Noun phrases in group I, where the Japanese contains a possessive
expression are directly translated at the beginning of the (sixth)
transfer stage.  In order to determine the antecedent of the pronoun
in noun phrases with the genitive reflexive construction {\it
  jibun-no}, {\bf ALT-J/E} uses a simple algorithm that identifies the
subject of the clause containing {\it jibun-no\/} `one's own' as the
antecedent after elided subjects have been
supplemented\footnote{Elided subjects are supplemented using
  information both from within the sentence being translated and from
  the surrounding paragraph \cite{Nakaiwa:1992}.}.  {\bf ALT-J/E} uses
a simple mapping of antecedent to pronoun based on the antecedent's
syntactic features of {\sc person}, {\sc gender} and the semantic
feature {\sc human} as shown in Figure~\ref{fig:def}.  If the
possessive pronoun itself appears in the subject, then the pronoun is
judged to be used deictically and is determined according to the
modality of the sentence, for example for declarative sentences the
pronoun is taken to refer to the speaker (giving {\it my\/}),
whereas for imperative or interrogative sentences it is taken to refer 
to the hearer (giving {\it your}).

\begin{figure}[htbp]
  \halfborder
    \begin{tabbing}
123\=123\=123\=123\=123\= \kill
      \> if the antecedent is a simple noun phrase \\
      \> \> if it is 1st per \\
      \> \> \> if it is Si ({\it I\/}) \\
      \> \> \> \> use \it my \\
      \> \> \> else if it is  Pl ({\it we}) \\
      \> \> \> \> use \it our \\
      \> \> if it is 2nd per  ({\it you}) \\ 
      \> \> \> use \it your \\
      \> \> else (treat it as 3rd per)\\
      \> \> \> if it is Si \\
      \> \> \> \> if it is the generic pronoun {\it one\/} \\
      \> \> \> \> \> use \it one's \\
      \> \> \> \> else if it is {\sc male} ({\it he}/{\it Mr Bond}) \\
      \> \> \> \> \> use \it his \\
      \> \> \> \> else if it is {\sc female} ({\it she}/{\it Ms Bond}) \\
      \> \> \> \> \> use \it her \\
      \> \> \> \> else if it is {\sc human} ({\it Dr Bond}) \\
      \> \> \> \> \> use \it their  (gender unknown)\\
      \> \> \> \> else (it is Si and non-human)  ({\it it}/{\it NTT}) \\
      \> \> \> \> \> use \it its \\
      \> \> \> else (it is  Pl) ({\it they\/}/{\it the G7 Nations}) \\
      \> \> \> \> use \it their \\
      \> else (the antecedent is a compound noun phrase) \\
      \> \> if one element is 1st per\\
      \> \> \> use \it our \\
      \> \> else if one element is 2nd per \\
      \> \> \> use \it your \\
      \> \> else (treat as 3rd per) \\
      \> \> \> use \it their 
    \end{tabbing}
  \caption{Determination of a possessive pronoun given its antecedent}
  \label{fig:def}
  \halfborder
\end{figure}
The translation rules for English expressions with obligatory
possessive pronouns identify the antecedent within the rule.  For
example the rule used in translating sentence~(\ref{sent:idiom}) can
be glossed as follows:

\noindent N1-{\it wa chie-o shiboru\/} `N1-{\sc top} knowledge-{\sc
  obj} wring' \\ 
$\rightarrow$ N1 {\it racks\/} N1'{\it s brains}\footnote{N1{\it 's
    \/} represents a possessive pronoun with N1 as its antecedent.}

 When the Japanese analysis stage has parsed the
sentence correctly and an appropriate pattern has been chosen in the
transfer stage then the correct possessive pronoun will be generated.

For the 105 sentences of group II where the translator uses an idiom
containing a possessive pronoun, the machine translation system does
not always choose the same idiom as the human translator.  In the
cases where the machine generates an idiom that does use a possessive
pronoun it is generated correctly.

\section{Generating possessive pronouns in noun phrases headed by
  trigger-nouns} 
\label{sec:narrow}

This section describes the proposed method for appropriately
generating possessive pronouns for noun phrases headed by trigger
nouns.  The discussion will be illustrated with examples of
translations from two versions of {\bf ALT-J/E}.  The original version
(hereafter the '93 version) does not use the proposed method for
generating possessive pronouns.  The version that uses the proposed
method will be referred to as the '94 version.\footnote{Translations
  made by {\bf ALT-J/E} before the proposed processing was included
  are marked ``MT-93''.  Translations done by the current version of
  {\bf ALT-J/E} which includes the proposed processing are marked
  ``MT-94''.}

The generation of possessive pronouns in noun phrases headed by
trigger-nouns occurs at the end of the transfer phrase.  We shall call
the pronouns generated for these noun phrases `default possessive
pronouns' because they are generated as a default, not as a result of
being explicitly indicated in the Japanese or in the translation
pattern.  

\begin{figure}[htbp]
  \halfborder
  \noindent   \begin{enumerate}
  \item A noun phrase that fulfills all of the following
    conditions will be generated with a default possessive pronoun
    with deictic reference determined by the modality of the
    sentence it appears in.
    \begin{enumerate}
    \item The noun phrase is headed by a trigger-noun that
      denotes {\sc kin} or {\sc body parts}
    \item The noun phrase is the subject of the sentence
    \item The noun phrase is referential
    \item The noun phrase has no other determiner
    \end{enumerate}
  \item A noun phrase that fulfills all of the following
    conditions will be generated with a default possessive pronoun
    whose antecedent is the subject of the sentence the noun
    phrase appears in.
    \begin{enumerate}
    \item The noun phrase is headed by a trigger-noun
    \item The noun phrase is not the subject of the sentence
    \item The noun phrase is referential
    \item The noun phrase has no other determiner
    \item The noun phrase is not the direct object of a verb of
      {\sc possession} or {\sc acquisition}
    \end{enumerate}
      \end{enumerate}
  \caption{Proposed method of generating possessive pronouns}
  \label{fig:proposed}
  \halfborder
\end{figure}

The proposed algorithm is outlined in Figure~\ref{fig:proposed}.
First, the noun phrase's referential property is determined as
described in Section~\ref{sec:gen}.  If the noun phrase's determiner
slot is already filled, then it cannot have a possessive pronoun.
Some of the ways that the determiner slot can be filled are described
in Section~\ref{sec:definite}.  Finally, if the noun phrase is headed
by a trigger-noun and is neither the subject of the sentence nor the
direct object of a verb with meaning {\sc possession} or {\sc
  acquisition}, then it will be generated with a possessive pronoun
whose antecedent is the subject of the sentence.  The significance of
the verb meaning is discussed in Section~\ref{sec:verb}.

The extra rules for noun phrases headed by trigger-nouns denoting {\sc
  kin} or {\sc body parts} are described in Section~\ref{sec:kin}.

\subsection{Effects of noun phrase referentiality}
\label{sec:gen}

The use of heuristic rules to determine the referentiality of noun
phrases in the machine translation system {\bf ALT-J/E} is discussed
in detail in \namecite{Bond:1995b}.  In the following discussion we
will assume that the referentiality of a noun phrase can be correctly
determined.

Consider the translation of {\it hana\/} `nose' in sentences
(\ref{sent:gen}) and (\ref{sent:ref}).

\begin{rei} 

\begin{tabular}{cllll}
(\sent{sent:gen})
    & Jap:   &  \it hana-wa &  \it kankakukikan & \it da  \\
    & Gloss: &   nose-{\sc top} &  sensory organ &  is  \\
    & Eng:   & \multicolumn{3}{l}{`The nose is a sensory organ'} \\
    & MT-93: & \multicolumn{3}{l}{\sf A nose is a sensory organ} \\
    & MT-94: & \multicolumn{3}{l}{\sf Noses are sensory organs}
\end{tabular}

\begin{tabular}{clll}
(\sent{sent:ref})
& Jap:   &  \it hana-ga & \it kayui \\
& Gloss: &   nose-{\sc top} &  itch \\
& Eng:   & \multicolumn{2}{l}{`My nose itches'} \\
& MT-93 & \multicolumn{2}{l}{\sf A nose itches} \\
& MT-94 & \multicolumn{2}{l}{\sf My nose itches} 
\end{tabular}

\end{rei}

In sentence (\ref{sent:gen}) the subject {\it nose\/} is determined by
the human translator to have generic reference and no possessive
pronoun is used.  In sentence (\ref{sent:ref}) the subject is
determined to refer to a specific person's nose, and so a possessive
pronoun is used.  In general, noun phrases with generic reference are
not modified by possessive pronouns.  Similarly, noun phrases used
ascriptively, to ascribe an attribute to another noun phrase, do not
use possessive pronouns, e.g. {\it That is a big nose!\/} Therefore,
we restrict the problem of determining when possessive phrases should
be used to referential noun phrases.

In sentence (\ref{sent:gen}), the '93 version does not differentiate
between generic and referential noun phrases.  By chance, the
translation given ({\it A nose is a sensory organ\/}) has a generic
interpretation so the translation is judged as correct.  In the '94
version, the system determines that the sentence is generic, because
it is stating a general truth, and thus the subject has generic
reference.  The judgment is done with the following rule: if the
semantic category of the subject of a copula is a child of the
semantic category of the object then the noun phrase in the subject
position has generic reference.  In this case, the semantic attributes
stored in the lexicon for {\it nose\/} and {\it sensory organ\/} are
{\sc nose} and {\sc organ} respectively, and the category {\sc nose}
is a child of {\sc organ}, that is, a {\sc nose is-a organ}.  Generic
noun phrases headed by countable nouns are translated as bare plurals
and are not candidates for the generation of default possessive
pronouns.  In sentence (\ref{sent:ref}), however, the subject is
determined to be referential.  Therefore, as {\it nose\/} is a
trigger-noun a possessive pronoun with deictic reference is generated.

\subsection{Filling the determiner slot}
\label{sec:definite}

Default possessive pronouns will not be generated if the determiner
slot has been filled.  The determiner slot can be filled by elements
directly translated from the Japanese: e.g. the demonstrative {\it
  kono\/} `this' fills the determiner slot as {\it kono\/} `this'.  It
can also be filled by the rules that generate definite and indefinite
articles: e.g., in the noun phrase in example~(\ref{sent:def}) even
though it is headed by the trigger-noun {\it saifu\/} `wallet' the
determiner slot is filled by the definite article so the noun phrase
is not a candidate for a default possessive pronoun.  In contrast, a
possessive pronoun is generated in example~(\ref{sent:notdef}) where
there is no definite article generated.  The rules that can fill the
determiner slot are too numerous to be enumerated here.

\begin{rei} 
\begin{tabular}{clllllll}
(\sent{sent:def})
    & Jap:   &  \sl watashi-wa & \sl kin\=o & \sl katta  & \sl saifu-o
    \\ 
    & & \sl nakushita \\
    & Gloss: &  I-{\sc top} &  yesterday & bought  & wallet-{\sc obj} \\
    & &  lost  \\
    & Eng:   & \multicolumn{5}{l}{`I lost the wallet I bought yesterday'}\\
    & MT-93: & \multicolumn{5}{l}{\sf I lost the wallet I bought yesterday}  \\
    & MT-94: & \multicolumn{5}{l}{= MT-93} \\
\end{tabular}
\end{rei}

\begin{rei} 
\begin{tabular}{clllll}
(\sent{sent:notdef})
    & Jap:   &  \sl watashi-wa & \sl saifu-o & \sl nakushita \\
    & Gloss: &   I-{\sc top}  & wallet-{\sc obj} &  lost  \\
    & Eng:   & \multicolumn{3}{l}{`I lost my  wallet'}\\
    & MT-93: & \multicolumn{3}{l}{\sf I lost  a  wallet}  \\
    & MT-94: & \multicolumn{3}{l}{\sf I lost  my  wallet} \\
\end{tabular}
\end{rei}

\subsection{Restrictions determined from the meanings of verbs}
\label{sec:verb}
Examining the test set showed that the meanings of verbs can be used
to determine whether a possessive pronoun should be generated or not
for noun phrases headed by trigger-nouns.  Noun phrases which are the
direct objects of verbs that express possession, such as {\it own,
  have\/} or {\it possess\/} and noun phrases that are the object of
verbs that express that the object has just been acquired, for
example, the direct object of {\it buy, acquire\/} or {\it steal\/}
are translated with an indefinite article rather than a possessive
pronoun even when headed by trigger-nouns.

Both these cases can be explained by considering the verb's meaning.
In the first case the verb itself shows that the subject is the
possessor of the object, so a possessive pronoun is not needed to show
the meaning.  If a possessive pronoun is used, it especially
emphasises the fact that the subject's referent possesses the referent
of the object.\footnote{A Japanese sentence that emphasises this
  possessive relationship would explicitly use {\it jibun-no}
  `self-{\sc gen}'.  In this case {\bf ALT-J/E} will generate a
  possessive pronoun.  For example, {\it jibun-no kutsushita-o
    motteimasu-ka\/} `self-{\sc gen} sock-{\sc obj} have-{\sc q}' {\it
    Do you have your own socks?}.} In the second case, in which the
subject `acquires' the object, the object is not `possessed' by the
subject until after the action described by the verb is completed, so
a possessive pronoun is not used.

{\bf ALT-J/E} classifies verb meanings using the system of 97 verbal
semantic attributes introduced in \namecite{Nakaiwa:1994}.  Verbs with
similar meanings share the same verbal semantic attributes which
allows a rule to be written as follows:

\begin{itemize}
\item If a noun phrase headed by a trigger-noun is the direct object
  of a verb of {\sc possession} or {\sc acquisition} then do not
  generate a possessive pronoun.\footnote{Furthermore if the noun
    phrase has no pre-determiner, determiner or post-determiner then
    maybe generate the determiner {\it some\/} (or {\it any\/}
    depending on the sentence aspect and noun phrase countability and
    number).}
\end{itemize}

This rule is exemplified in sentence~(\ref{sent:some}).  If this rule
were not implemented then because {\it kuruma\/} `car' is a
trigger-noun the sentence would have been incorrectly translated as
`Do you have your car?' which introduces an emphasis that the
original Japanese lacks.

%
%
\begin{rei} 
\begin{tabular}{clll}
(\sent{sent:some})
    & Jap:   &  \sl kuruma-o & \sl motteimasu-ka  \\
    & Gloss: &  car-{\sc obj} &  have-{\sc Q}  \\
    & Eng:   & \multicolumn{2}{l}{`Do you have a car?'} \\
    & MT-93: & \multicolumn{2}{l}{\tt Do you have  a car?} \\
    & MT-94$'$: & \multicolumn{2}{l}{\tt Do you have your  car?} \\
    & MT-94: & \multicolumn{2}{l}{\tt Do you have a  car?}
\end{tabular}

\end{rei}

\subsection{{\sc kin} and {\sc body parts}}
\label{sec:kin}

In the test set, noun phrases denoting {\sc kin} or {\sc body parts}
are modified by possessive pronouns used deictically when they are the
subject of the sentence.  Therefore, the pronoun is determined
according to the modality of the sentence: e.g. for declarative
sentences the pronoun is first person singular (giving {\it my\/}),
whereas for imperative or interrogative sentences it is the second
person (giving {\it your\/}).

Two special cases were identified.  Nouns which explicitly denote {\sc
  parents} or {\sc children}\footnote{That is nouns such as {\it
    child\/} but not nouns such as {\it son}.} are only translated
with possessive pronouns if they appear together in the same sentence.
In this case, they are translated as though they are related to each
other but not to the speaker.  Therefore the following special rule
has been implemented: Only generate a possessive pronoun for trigger
nouns which explicitly denote {\sc parents} or {\sc children} if a
sentence contains one of each category, in which case the first to
appear is the antecedent of the second to appear.

The second special case was for sentences with compound subjects that
include nouns that denote {\sc kin}.  For example, if the subject is
{\it me and my spouse\/} and the person in the noun phrase in question
is a member of the family other than `our' children (or grandchildren)
then they will normally be either related to {\it me\/} or to {\it my
  spouse}, but not both, therefore they will be modified by {\it my\/}
rather than {\it our}: e.g. {\it My wife and I gave my sister a book\/}
but {\it My wife and I gave our child a book}.  Similarly siblings
will not normally have children or grandchildren in common so {\it
  my\/} will be used for children and grandchildren: {\it My sister
  and I gave our mother a book\/} but {\it My sister and I gave my child
  a book}.  These rules have not yet been implemented.

\section{Results}
\label{sec:results}

A preliminary evaluation of {\bf ALT-J/E}'s generation of possessive
pronouns was conducted on the test set of 6,200 sentences described in
Section~\ref{sec:diff}.  All save two of the 168 noun phrases in group
I,in which the original Japanese contained an explicit possessive
expression, are translated correctly. Two of the 25 sentences
containing {\it jibun-no\/} in the test set were translated
incorrectly.  Both of the errors were caused by the subject being
incorrectly identified in embedded sentences.  For the 105 noun
phrases in group II, in which a possessive pronoun is required by an
English expression in the human's translation but there is no
possessive expression in the Japanese, {\bf ALT-J/E} did not always
select the same expression as the human translator.  When {\bf
  ALT-J/E} selected an expression that requires a possessive pronoun,
such as {\it to rack one's brains\/} in sentence~(\ref{sent:idiom}) or
{\it to wash one's hands\/} in sentence~(\ref{sent:idiom2}), it was
generated correctly.

\begin{rei}
\begin{tabular}{cllll}
(\sent{sent:idiom2})
& Jap:   &  \it kare-wa & \it kono-shigoto-kara & \it te-o \\
& Gloss: &   he-{\sc top}  &  this work from & hands-{\sc obj} \\
&        & \it hiita \\
&        & pulled \\
& Eng:   & \multicolumn{3}{l}{`He washed his hands of this work'}
\end{tabular}
\end{rei}

There were 825 noun phrases in the test set headed by trigger-nouns.
In 9\% of the noun phrases (73), there were errors in the analysis or
transfer stages which made evaluation of the appropriateness of the
possessive pronoun impossible. The results of the generation of the
remaining 752 noun phrases are given in Table~\ref{tab:results}.

\begin{table*}
  \border
  \begin{center}
    \begin{tabular}{|l|l|rr|rr|l|} \hline
      Result & \multicolumn{1}{c|}{Possessive}  &
      \multicolumn{2}{c|}{MT-93} & \multicolumn{2}{c|}{MT-94} & \multicolumn{1}{c|}{Example}\\ 
      & \multicolumn{1}{c|}{pronoun}     & NPs & \multicolumn{1}{c|}{Percentage} & NPs &
     \multicolumn{1}{c|}{Percentage} & \\ \hline
      Good & Not generated & 429 & 57\% & 346  & 46\% & I hit him in {\it the\/} face \\
      & Generated     &   0 & 0\%  & 263  & 35\% &  I hid {\it my\/} face\\ \cline{2-7}
      & \bf --- Total --- & 429 &  57\% & 609 & 81\% &  \\ \hline
      Bad  & Not generated & 323 & 43\% &  60 & 8\% & * I scratched {\it a\/} face \\
      & Generated     &   0 & 0\%  &  83 & 11\% & * I lost {\it my\/} face \\ \cline{2-7}
      & \bf --- Total ---    &  323 & 43\% & 143 & 19\% &  \\  \hline 
    \end{tabular}
  \end{center}
  \caption{Results of the generation of noun phrases headed by
    trigger-nouns (Total 752 noun phrases).} 
  \label{tab:results}
  \border
\end{table*}

The evaluation was conducted by comparing the machine generated
translation of the noun phrases headed by trigger-nouns with the human
translations.  A machine generated possessive pronoun is judged to be
appropriate if it also appears in one or more of the human
translations.  If a pronoun is generated that does not appear in the
human translations, it is judged to be not appropriate.\footnote{In
  25\% of the noun phrases in which the human translation had no
  possessive pronoun but {\bf ALT-J/E} generated one the developers
  judge that generating a possessive pronoun gives an interpretation
  as appropriate as the human translation.  For the purpose of this
  evaluation however they are treated as incorrect.} 429 (57\%) of the
noun phrases headed by trigger-nouns do not require a possessive
pronoun to be generated by the proposed method.  For example the noun
phrase phrase is non-referential, or the determiner slot is already
filled, or the noun phrases was dominated by a verb of {\sc
  possession} or {\sc acquisition}.  These noun phrases are all
translated correctly by the `93 version as it has no special
processing for generating possessive pronouns.  It fails, however, to
generate possessive pronouns when they are judged as necessary in the
remaining 323 noun phrases (43\%).  Thus the accuracy of the '93
version (the number judged correct over the total number) is only 57\%
(429/752)\footnote{Note that the original method only actually handles
  52\% of the total noun phrases correctly, however for the purpose of
  evaluating the algorithm we are ignoring the 73 noun phrases where
  the errors in the analysis and transfer stages are so great as to
  make the output unevaluable.}.

In the '94 version, using the proposed method, noun phrases are
generated when wanted 80\% of the time (the number of noun phrases
with appropriate possessive pronouns generated (263) over the number
of noun phrases where a possessive pronoun was judged appropriate
(323)).  The errors caused by not generating the appropriate pronoun
are mainly due to errors in the parse selected in the analysis stage
and conflicts with other rules.  We estimate that overall improvements 
in the parsing and transfer stages can solve these problems for
30 of the noun phrases considered.  Thus the estimated potential
success rate is 91\% (293/323).   

The proposed method, however, introduces a new source of errors,
over-generation of possessive pronouns.  Possessive pronouns are
inappropriately generated for 83 noun phrases headed by trigger-nouns
(11\% of the total number).  Two solutions are proposed.  First, to
improve the processing that determines the noun phrase referentiality
and definiteness, this would block possessive pronouns from being
generated by filling the determiner slot with a more appropriate
determiner.  Second, to introduce explicit semantic constraints (e.g.:
only generate a possessive pronoun for trigger-nouns that denote {\sc
  clothing} in the object position if the subject denotes a {\sc
  human}), these would stop pronouns from being generated
unnecessarily.  We estimate that a combination of these solutions can
reduce the over-generation to around 45 noun phrases (6\%).  To reduce
the errors beyond this, we would require a discourse analysis capable
of actually determining explicit possessive relationships within a
local world model. Until such an analysis becomes feasible, some
over-generation is inevitable with the proposed method.  We make it
easier to correct for this during post-editing by tagging possessive
pronouns generated by the proposed method as being less reliable than
possessive pronouns generated from directly from possessive
expressions in the source text or transfer patterns.  The tagged
pronouns can then be marked when presented to a post editor (for
example in a different colour or font) for special attention.

The accuracy of the '94 version (the number judged correct over the
total number) is 81\% (609/752), an improvement of 24\%.  The
precision of for the new method (the number judged correct out of the
total number generated by the proposed method) is 88\% (609/692).  The
accuracies and precisions achieved by the '93 version (which does not
use the proposed method) and the '94 method (which uses the proposed
method) for the generation of possessive pronouns in noun phrases
headed by trigger-nouns are summarized in Table~\ref{tab:overall}.

\begin{table}[htbp]
  \halfborder
  \begin{center}
    \begin{tabular}{lcr}
      Result   & \multicolumn{1}{c}{MT-93} & \multicolumn{1}{c}{MT-94}\\ \hline
      Accuracy &  57\% &  81\% \\
      Precision & --- &  88\% 
    \end{tabular}
  \end{center}
  \caption{Overall evaluation of proposed method.}
  \label{tab:overall}
  \halfborder
\end{table}

\section{Conclusion}
\label{sec:conc}

In order to examine when possessive pronouns should be generated when
translating between Japanese and English 6,200 Japanese sentences with
English translations were examined.  657 examples of noun phrases
containing possessive pronouns were found in the human translations.
The existing algorithms used by the Japanese-to-English machine
translation system {\bf ALT-J/E} were sufficient for 46\% of the noun
phrases. A heuristic method for appropriately generating possessive
pronouns for the remaining 54\% was proposed.  The method uses cue
words we call trigger-nouns, along with contextual information about
noun phrase referentiality and the subject and main verb of the
sentence that the noun phrase appears in.  The proposed method was
implemented in {\bf ALT-J/E}.  It increased the number of noun phrases
with appropriate possessive pronouns generated by 263 to 609, but at
the cost of generating 83 noun phrases with inappropriate possessive
pronouns.  We intend to increase the number of appropriate possessive
pronouns generated by resolving rule conflicts and to reduce the
number of inappropriate possessive pronouns generated by adding more
semantic constraints.


\section*{Acknowledgments}

We would like to thank Tsuneko Nakazawa for her comprehensive
criticism and advice; the reviewer, Graham, Monique and Mitsuyo
Bond for their comments and suggestions; and Toshiaki Nebashi, Kazuya
Fukamachi and Yoshitake Ichii for their invaluable help in
implementing the processing described here.


\end{document}